\pdfoutput=1
\documentclass[11pt]{article}

\usepackage[]{ACL2023}

\usepackage{times}
\usepackage{latexsym}

\usepackage[T1]{fontenc}
\usepackage[utf8]{inputenc}

\usepackage{microtype}

\usepackage{inconsolata}

\usepackage[utf8]{inputenc} % allow utf-8 input
\usepackage[T1]{fontenc}    % use 8-bit T1 fonts
\usepackage{hyperref}       % hyperlinks
\usepackage{url}            % simple URL typesetting
\usepackage{booktabs}       % professional-quality tables
\usepackage{amsfonts}       % blackboard math symbols
\usepackage{nicefrac}       % compact symbols for 1/2, etc.
\usepackage{microtype}      % microtypography
\usepackage{xcolor}         % colors
\usepackage{graphicx}
\usepackage{amsmath}
\usepackage{bbm}
\usepackage{cleveref}
\usepackage{amsthm}
\theoremstyle{definition}

\theoremstyle{lemma}

\theoremstyle{remark}

\usepackage{amssymb}
\usepackage{tcolorbox,mathtools,biolinum}
\usepackage{hyperref}
\tcbuselibrary{skins,theorems}
\usepackage{amsthm}

\usepackage{xcolor}         % colors
\usepackage{amssymb}% http://ctan.org/pkg/amssymb
\usepackage{pifont}% http://ctan.org/pkg/pifont
\usepackage{url}

\usepackage{color}
\NewDocumentCommand{\vicki}{mO{}}{\textcolor{blue}{\textsuperscript{\textit{Vicki}}\textsf{\textbf{\small[#1]}}}}

\title{C-PMI: Conditional Pointwise Mutual Information for \\Turn-level Dialogue Evaluation  }

\author{
Liliang Ren\thanks{~~Equal contribution.},
~Mankeerat Sidhu\footnotemark[1],
~Qi Zeng, ~Revanth Gangi Reddy, \\
\textbf{Heng Ji, ~ChengXiang Zhai}\\
University of Illinois Urbana-Champaign \\
\texttt{\{liliang3, mssidhu2, qizeng2, revanth3, hengji, czhai\}@illinois.edu }
}

\begin{document}
\maketitle
\begin{abstract}

Existing reference-free turn-level evaluation metrics for chatbots inadequately capture the interaction between the user and the system.
Consequently, they often correlate poorly with human evaluations.
To address this issue, we propose a novel model-agnostic approach that leverages Conditional Pointwise Mutual Information (C-PMI) to measure the turn-level interaction between the system and the user based on a given evaluation dimension.
Experimental results on the widely used FED dialogue evaluation dataset demonstrate that our approach significantly improves the correlation with human judgment compared with existing evaluation systems. 
By replacing the negative log-likelihood-based scorer with our proposed C-PMI scorer, we achieve a relative 62.6\% higher Spearman correlation on average for the FED evaluation metric. Our code is publicly available at \url{https://github.com/renll/C-PMI}.

\end{abstract}
\section{Introduction}

% \vicki{@Mnakeerat: the background part is too long}
% The rapid advancements in social bot technology, including Alexa, Google Assistant, and ChatGPT, have emphasized the importance of automatic evaluation of dialogue systems, particularly in the context of Large Language Models (LLMs) such as GPT-4. 
Evaluating dialogues is a multi-faceted task that demands consideration of diverse dimensions, which distinguishes it from the evaluation of task-oriented dialogue systems. Traditional n-gram-based evaluation metrics, such as ROUGE~\cite{lin-2004-rouge} and BLEU~\cite{papineni-etal-2002-bleu}, demonstrate weak correlation with human-annotated judgments due to the broad spectrum of potential responses in dialogues. As a result, researchers often resort to human evaluations to ascertain the quality and effectiveness of their generated system responses, especially for knowledge-guided dialog systems~\cite{dialog2022,FungEACL2023,LaiEACL2023}.

Substantial research has been conducted on automatic evaluation metrics for dialogue~\cite{yeh-etal-2021-comprehensive}. These metrics can be classified into reference-based and reference-free categories. Reference-based metrics, which depend on comparing the system response to a human-written reference response, are generally inadequate for dialogue evaluation due to the inherent one-to-many nature of dialogues. The reference-free metric instead uses a computational model to generate a score for the system response with a given context.
% \vicki{@mankeerat: introduce reference-free methods here, not learning-based methods}
% Learning-based metrics, such as ADEM~\cite{lowe-etal-2017-towards} and RUBER~\cite{DBLP:conf/aaai/TaoMZY18}, employ Recurrent Neural Networks to directly predict the quality of system responses. BERT-RUBER~\cite{ghazarian-etal-2019-better} advances RUBER's unreferenced metric using BERT.

Early models predominantly focus on a limited set of general features of dialogue generation quality, such as context coherency and fluency. Subsequent evaluation metrics investigated additional dimensions, such as USL-H~\cite{phy-etal-2020-deconstruct}, which combines relevance evaluation with fact-to-response selection. Holistic-eval ~\cite{pang-etal-2020-towards} assesses content coherence, language fluency, self-consistency, and semantic appropriateness. D-Score ~\cite{DBLP:journals/taslp/ZhangLDL21} and Predictive Engage ~\cite{ghazarian2020predictive} introduce response diversity and engagement scores. The recent FED \cite{fed} metric encompasses 18 turn-level and dialogue-level metrics, including interestingness, likeability, and response flexibility. 
%Many of these models are unreferenced and do not require training, making them suitable for response selection. FED utilizes DialoGPT, DialogueRPT fine-tunes GPT-2 for response quality prediction, and GPTScore employs GPT-3, FLAN-T5, and other pretrained LLMs for dialogue evaluation. 
However, all of these methods do not model the interaction between the turn-level response and the dialogue history and regard them as an integrated context for score calculation.

%we concentrate on devising and validating a novel algorithm that harnesses an LLM to generate a score, which can be extended to encompass eight turn-level evaluation metrics and a combined turn-level score. Our primary methodological innovation is based on
In this paper, we focus on directly modeling user-system interactions through the lens of Mutual Information \cite{shannon,ghassami2017interaction} and propose a novel scorer based on Conditional Pointwise Mutual Information (C-PMI), which effectively captures the turn-level interactions between the system and user with respect to a given hypothesis. 
%The C-PMI-based turn-level metric provides a model-agnostic approach that serves as a generalized alternative to the Negative Log-Likelihood (NLL) based scoring method. 
We demonstrate that our approach results in a reference-free, training-free, automatic turn-level dialogue evaluation that significantly outperforms state-of-the-art methods with a comparable number of model parameters. Our contributions in this work are three-fold:
\begin{itemize}
    \item  A novel dialogue evaluation metric based on Conditional Pointwise Mutual Information (C-PMI) that effectively captures turn-level interactions between the system and user with respect to a given hypothesis.
    \item An unreferenced, training-free, automatic turn-level dialogue evaluation that significantly outperforms state-of-the-art methods with a comparable number of model parameters. 
    \item A model-agnostic approach that can be served as a generalized alternative to the Negative Log-Likelihood (NLL) based evaluation metrics when interactions between previous turns need to be considered.
\end{itemize}

%In this paper, we will be focusing specifically on the Interestingness score of a dialogue with a novel algorithm and leveraging GPT-3 to generate a score. We also additionally release a total of 8 metrics and a combined turn-level evaluation metric. Interestingness alongside relevance and engagement has shown to be the biggest contributor in making a good dialogue (insert citation), making almost 20\% of total dialogue evaluation out of all other dimensions. The contributions of this paper are 1) a strongly-correlated, unsupervised, and reference-free metric for evaluating turn level interestingness in open-domain dialogue systems, 2) a novel GPT-3 leveraged algorithm  (describe further @liliang) and 3) TTIM (Total turn-level Informative metric) presenting 8 dimensions including interestingness, fluency, engagingness, specificity, relevance, correctness/appropriateness and understandability based on our algorithm.

\section{Related Work}

Developing automatic evaluation metrics for dialog is challenging for several reasons:
1) Dialogues often have a one-to-many nature, rendering word-overlap metrics ineffective. To address this issue, metrics should be designed to be reference-free.  
2) Given the limitless nature of conversation topics in open-domain dialogues, the dialogue evaluation metrics are expected to understand the semantic meaning of both the dialogue context and the generated responses. This necessitates a metric that can leverage pre-trained large language models and self-supervised training objectives.
3) Training dialogue evaluation metrics solely on labeled data can significantly restrict the metric's range, risking over-fitting to the training data in terms of conversation topics and response generation models. As such, recent metrics have started to incorporate self-supervised training objectives designed to capture various aspects of a dialogue, such as relevance, fluency, and interestingness among others.

Given the aforementioned challenges, large language models have become an integral part of dialogue evaluation.
DialogRPT~\cite{gao2020dialogrpt} employs an extended GPT-2 model trained on 147 million conversation-like interactions from Reddit. 
USR~\cite{mehri-eskenazi-2020-usr} is an unsupervised, reference-free tool that takes advantage of the RoBERTa~\cite{DBLP:journals/corr/abs-1907-11692} model.
USR employs a dialogue retrieval metric for assessing dialogue, where the metric is trained to differentiate between a ground truth response and a randomly sampled response. 
The FED metric~\cite{fed} utilizes DialoGPT~\cite{zhang2019dialogpt} due to its capacity for capturing knowledge, specifically within the context of conversations. 
It ignores the interaction between the user and the system and consider the dialogue history and the system response as an integral context, while our method explicitly captures such interaction through conditional mutual information.
% \section{Background}

% Past works ignore modeling the turn-level interaction between the system and the user given a hypothesis. e.g. NLL based metric used by FED, GPT score, UniEval, etc.

\section{Background}
FED~\cite{fed} measures eighteen fine-grained qualities of dialogue without requiring comparison to a reference response or training data with ground-truth human ratings.
The method leverages DialoGPT and uses the follow-up hypotheses as a means of evaluation, based on the assumption that the language model has learned to accurately measure the likelihood of the input sequence.
Given a dialog context $c$, a system response $r$, and a scorer $\mathcal{L}$ that computes the average Negative Log-Likelihood (NLL) of a sequence with a language model $\theta$, the predicted score for a pair of positive and negative hypotheses $(p_i,n_i)$ is calculated as,
% \textbf{FED}~\cite{fed} is an automatic evaluation metric not requiring a comparison to a reference response, measures eighteen fine-grained qualities of dialogue and does not require training data with ground-truth human ratings. The method leverages DialoGPT, relying on the intuition that DialoGPT has implicitly learned to reveal quality data, and use of ‘follow-up’ utterances as a means of evaluation. In FED, given a dialog context c, a system response r and a function D that computes the log-likelihood of DialoGPT generating a particular response, the predicted score for a dialogue quality is calculated as:
$$
    \sum_{i=1}^{|n|} \mathcal{L}\left(\{c,r, n_i\}, \theta \right)-\sum_{i=1}^{|p|} \mathcal{L} \left(\{c,r, p_i\}, \theta \right),
$$
where $\{a, b\}$ means text $b$ is appended to text $a$,and for each of the evaluation dimensions, $|p|$ and $|n|$ number of positive and negative hypothetical sentences are respectively pre-defined and used for reducing evaluation variance.
For example, given a combined history $\{c,r\}$, the response is regarded as more interesting if the probability of DialoGPT generating a positive hypothesis (e.g., "That's really interesting!") is greater than the probability of it generating a negative one (e.g., "That's really boring.").

\section{Conditional Pointwise Mutual Information based Turn-level Metric}

For each of the dialogue turn $t$, our Pointwise Mutual Information (PMI) based metric is considering the dependencies between the following three random variables: the full dialogue history $\mathbf{r}_t = \{u_0, x_0, u_1, x_1, ..., u_t \}\sim R$ (where $u_t$ is the user utterance), 
the system response $x_t \sim X$ and a hypothesis $h\sim H$. Ideally, we want to know how much correlation between the dialogue history and the system response causes the hypothesis to be a plausible entailment of the combined history, $\{\mathbf{r}_t, x_t\}$. We measure such correlation by calculating the Conditional Mutual Information (CMI) between the response and the history with a given hypothesis, \emph{i.e.},
 \begin{align*}
      I(R,X|H) &= \mathbb{E}_{R,X,H} [\log \frac{p( \mathbf{r}_t, x_t | h ) }{ p(\mathbf{r}_t | h  )p(x_t |h )}] \\
      & = \mathbb{E}_{R,X,H} [\log \frac{p( \mathbf{r}_t, x_t ,h ) p( h ) }{ p(\mathbf{r}_t, h  )p(x_t, h )}].
 \end{align*}
 % where $\{a, b\}$ means text $b$ is appended to text $a$. 
 Intuitively, if $I(R,X|H)$ is large, the hypothesis is less likely to be caused by the interaction (\emph{i.e.}, the shared information) between $R$ and $X$.

Since sampling the history on a turn-by-turn basis needs exponentially increasing computation, an accurate estimation of the CMI between these random variables is intractable. Therefore, we propose to measure the CMI by calculating the pointwise mutual information contained between the observed dialogue history and the system response when the hypothesis is appended to the combined history.
 Formally, we define our Conditional PMI (C-PMI) score between the observed dialogue history, the system response, and the hypothesis as follows,
$$
    \text{C-PMI}( \mathbf{r}_t, x_t | h) = \log \frac{p(\mathbf{r}_t,x_t, h ) p(h)}{ p(\mathbf{r}_t, h)p(x_t, h)}.
$$
In practice, we estimate the probability of each sequence using the averaged Log-Likelihood (LL) obtained from a language model $P_\theta$, \emph{i.e.},
 $$
 \text{LL}(\mathbf{s}) = \frac{1}{n} \sum_{i=1}^{n} \log P_\theta(s_i |\mathbf{s}_{<i}),
$$
and our score is then computed as,
\begin{align*}
      \text{C-PMI}( \mathbf{r}_t, x_t | h) = & ~\text{LL}(\mathbf{r}_t,x_t, h ) + \text{LL}(h)\\
        &- \text{LL}(\mathbf{r}_t, h)- \text{LL}(x_t, h),
\end{align*}
which can be efficiently implemented using the modern deep learning framework. To retain the symmetric property of the mutual information, we also define a symmetric version of our score, C-PMI-SYM, by interchanging the response and the dialogue history, \emph{i.e.},
 \begin{align*}
         \text{C-PMI-SYM}( \mathbf{r}_t, x_t | h)= &\frac{1}{2}(\text{C-PMI}(\mathbf{r}_t, x_t | h) \\
    & +\text{C-PMI}( x_t ,\mathbf{r}_t| h) ).
 \end{align*}
For integrating our scorer with the existing evaluation system such as FED, we simply replace its NLL scoring function with our C-PMI scorer, and follow the original pipeline to get the final score for each of the data samples.

%such as PyTorch \cite{pytorch}.

% \begin{itemize}
%     \item LLM (explain) (Mankeerat TODO)
%     \item Algorithm (Liliang TODO)
%     \item Evaluating with GPT-3 (Liliang TODO)
% \end{itemize}

% Problem Formulation:
% User utterance at turn i: 
% $u_i$
% System response at turn i: 
% $x_i$
% Dialogue history: 
% 
% Dialogue System, DS:
% A conditional language model estimates $P(x_i |R_i )$

% Context: $u$

% Response with hypothesis: $x,h$

% %PMI = log(P([u,x])) - log(P(x)) -log(P(u))
% Perplexity based MI score :

% $\exp(-(\log(P(u,x,h))/N_a-\log(P(u))/N_u-\log(P(x,h))/(N_x+N_h)) )$ 

\section{Experiments}

\begin{table*}[htp]\centering
\small 
\begin{tabular}{lcccccccccc}
\toprule
\bf Metrics & \bf Interesting & \bf Fluent
 &\bf Engaging & \bf Specific & \bf Relevant & \bf Correct & \bf Appro. & \bf Und.  & \bf Avg. \\\midrule

% BARTSCORE+CNN & -3.3 & 17.2 & 1.1 & -7.9 & 10.0 & 1.8 & 18.8 & 8.1 & 5.7 \\
%  BARTSCORE+CNN+Para & -10.1 & 28.4 & -2.5 & -16.2 & 19.4 & 12.4 & 26.1 & 4.5 & 7.7 \\
\multicolumn{10}{l}{\emph{Supervised with Human Evaluations}}\\
\midrule
DynaEval &  32.7 &  17.1 & 30.0 & 34.6 & 26.3 & 24.2 &  20.2 &  20.0 &  25.6 \\
\midrule
\emph{Unsupervised}\\
\midrule
BARTSCORE & 15.9 & 14.0 & 22.6 & 8.3 & 11.9 & 7.6 & 10.0 & \bf 12.0 &  12.8 \\
FED & 32.4 & -13.4 & 24.0 & 14.1 & \bf 19.9 & \bf 26.2 & -9.4 & 1.3 & 11.9 \\
FED$^*$ & 32.5 &\emph{1.5} &17.6 & 23.0 & \underline{13.4} &15.9 &\emph{7.7} & \emph{6.0} &14.7 \\
FED + C-PMI-SYM & \bf 48.4 & \underline{16.6} & \underline{36.9} & \underline{28.0} & 10.5 & 14.8 & \underline{17.9} & 10.7 & \underline{23.0} \\
FED + C-PMI & \underline{48.2} & \bf 17.6 & \bf 37.0 & \bf 28.7 & 12.8 & \underline{17.6} & \bf 18.1 & \underline{11.1} & \bf 23.9 \\
%   \hline
% Alexaranker & 26.1 & 14.6 & 26.6 & 23.9 & 28.7 & 29.6 & 25.6 & 19.9 & 33.3 \\
\bottomrule
\end{tabular}
\caption{The Spearman correlations with human judgment on the FED Turn-level dataset. Italicized values indicate that they are not statistically significant (p > 0.05). We include the results from the supervised metric to showcase the power of our method. For the unsupervised metrics, the highest correlation is shown in bold and the second highest is underlined. $^*$ indicates our reimplementation. The results for DynaEval, BARTSCORE, and FED are from \citet{fu2023gptscore}. Appro. and Und. are respectively the abbreviations of the evaluation dimensions: Semantically Appropriate and Understandable.}\label{tab:results}
\end{table*}

\subsection{Dataset}

We evaluate our model on the turn-level annotated subset of the FED~\cite{fed} dataset. This subset consists of 455 data samples, each of which includes a dialog context, a system response, and eight human-annotated turn-level labels: Interesting, Fluent, Engaging, Specific,  Relevant, Correct,  Appropriate, and Understandable.
The annotations are obtained through a survey with the options of No, Somewhat, Yes, or N/A.
An additional overall impression label is measured using a five-point Likert Scale. 
The FED dataset is proposed to evaluate metrics as it is annotated with human quality judgments with conversations from Meena and Mitsuku bots ~\cite{DBLP:journals/corr/abs-2001-09977}.

% We use the FED \cite{fed} dataset to evaluate our model against other baselines and human judgments. The FED dataset consists of 3348 turn-level and 1364 dialog-level data points, including 455 data points with eight turn-level labels. The dataset's purpose is to evaluate metrics as it is annotated with human quality judgments with conversations from Meena and Mitsuku bots ~\cite{DBLP:journals/corr/abs-2001-09977}. The FED dataset also provides additional 18 dimensions including eight turn-level metrics which we also use to evaluate our total score. The turn level annotations were made using a survey with the options No, Somewhat, Yes, N/A and the overall impression was measured on a five-point Likert Scale. 

%The dataset stats are provided in Figure x.

% \begin{table}[htb]
% \small
% \centering
% \begin{tabular}{lcc}
% \toprule[1pt]
% aaa & bbb & ccc \\
% \hline
% aaa & bbb & ccc \\
% \bottomrule[1pt]
% \end{tabular}
% \caption{
% Dataset statistics 
% }
% \label{table:dataset}
% \end{table}

\subsection{Baseline Metrics}

We primarily compare our proposed reference-free and unsupervised metric with FED, but other baselines are also included as follows.

% We are proposing a reference-free and training-free metric and as to the best of our knowledge only the FED metric, as noted by \citet{yeh-etal-2021-comprehensive}, is the other untrained and reference-free metric. A whole list of previous metrics and whether they use training data or references can be found in Appendix A. Alongside FED, we also compare our results with some other metrics which use training data to show the strength of the algorithm proposed. 

%RUBER - It is a referenced metric and Unreferenced metric Blended Evaluation Routine which takes into consideration both the groundtruth reply and the query. It is learnable and its training does not require labels of human satisfaction. It was one of the first reference free metric achieving high correlation with human judgement due to its neural network based scorer measuringt he relatedness between the generated reply and its query. 
%BERT-RUBER - A RUBER based metric however replacing he static word embeddings with pretrained BERT contexualized embeddings trained on DailyDialogue dataset. 
%USR - It stands for an UnSupervised Reference-Free evaluation metric that leverages a pre-trained LM,  RoBERTa to measure several desirable qualities in a dialogue. The fine tuned model is then used to estimate the likelihood of a response using their MLM, predicting a probability distribution of a masked word is able to compute log likelihood of a response and showcases higher than before correlation with human judgement. 

\textbf{BARTScore}~\cite{NEURIPS2021_e4d2b6e6} is a text-scoring model based on BART ~\cite{lewis-etal-2020-bart} and does not requiring any fine-tuning.  
BARTScore calculates the weighted log probability of text $\mathbf{y}$ given text $\mathbf{x}$:
% The general form of BARTScore is:
$$
    \text { BARTSCORE }=\sum_{t=1}^m \omega_t \log P_\theta\left(\mathbf{y}_t \mid \mathbf{y}_{<t}, \mathbf{x}\right),
$$
where the weighted sum of the log probability of one text $\mathbf{y}$ given the other text $\mathbf{x}$ is used for scoring.
% BARTScore can also be used in different evaluation scenarios such as faithfulness, precision, recall and F score. 

% \paragraph{BARTScore+CNN} \cite{NEURIPS2021_e4d2b6e6} is based on BART and BARTScore finetuned on the CNNDM dataset ~\cite{NIPS2015_afdec700}.

% \paragraph{BARTScore+CNN+Para} \cite{NEURIPS2021_e4d2b6e6} is based on BART and BARTScore finetuned on the CNNDM dataset and Paraphrase2.0 ~\cite{hu-etal-2019-large}.

\textbf{DynaEval}~\cite{zhang-etal-2021-DynaEval} is an automatic evaluation framework for dialogue response generation tasks, designed to evaluate both turn-level and dialogue-level. The framework utilizes structured graph representations of dialogues and is trained on datasets that contain ground-truth human ratings. 

% \textbf{DynaEval}~\cite{zhang-etal-2021-DynaEval} is a unified automatic evaluation framework for dialogue response generation tasks on both turn level and dialogue level. It is predominantly motivated by dialogue coherence modeling and takes advantage of the structured graph representation of dialogues. This approach is trained on the datasets with ground-truth human ratings.

% Since our primary baseline, FED, uses DialoGPT as a model to generate the quality of the response, we also show our results using DialoGPT, keeping in mind the fairness of algorithm evaluation. 

% Method comparison - (On Overleaf) (LILIANG TODO - writing) (MANKEERAT TODO - add chart and finalize running all baselines and results)

\subsection{Implementation Details}

We follow the data pre-processing procedure as used by \citet{yeh-etal-2021-comprehensive} for the FED dataset
% \footnote{\url{https://github.com/exe1023/DialEvalMetrics/blob/main/data/fed_data/data_loader.py}}
, and modify the scorer function as in the original FED repository. 
% \footnote{\url{https://github.com/Shikib/fed/blob/fd498618c669f590cb5d78e6b55a70240e967925/fed.py#L29}}. 
Following \citet{yeh-etal-2021-comprehensive}, we use a special ``<|endoftext|>'' token to connect each turn of the system responses and the user utterances for constructing a full sequence. The sequence is then fed into the \emph{DialoGPT-large} language model to obtain the log-likelihood for calculating both the FED score and our C-PMI score.

\section{Results \& Analysis}

Table~\ref{tab:results} shows that our proposed metrics, FED+C-PMI-SYM and FED+C-PMI, outperform other methods in most of the evaluation dimensions, and is comparable to DynaEval which requires training on the evaluation dataset. Both FED+C-PMI-SYM and FED+C-PMI show substantial improvements in Interesting, Engaging, Specific, Semantically Appropriate and the Understandable dimensions compared to our re-implemented FED metric. Notably, our metric even substantially outperforms DynaEval on the Interesting and the Engaging dimensions which conceptually needs an accurate measure of the interaction between the user and the system. This demonstrates the effectiveness of our approach in capturing turn-level interactions.

The performance of FED+C-PMI-SYM and FED+C-PMI is quite similar across most dimensions. However, FED+C-PMI shows slightly better performance in the Relevant, Correct, and Understandable dimensions, suggesting that the asymmetrical variant of the C-PMI calculation might provide more accurate evaluation scores in certain cases. We suspect that this is because interchanging the positions of the response and the dialogue history results in unnatural dialogue, which leads to worse probability estimation from the language models.

The results indicate that the proposed C-PMI-based turn-level metrics are capable of providing a more accurate evaluation of dialogue system responses compared to existing state-of-the-art methods. Moreover, our metric is unreferenced and training-free, which makes it particularly suitable for practical applications, such as responses selection and re-ranking.

\section{Conclusion}

In this paper, we introduce a novel dialogue evaluation metric based on Conditional Pointwise Mutual Information (C-PMI) that captures turn-level interactions between the system and user across various evaluation dimensions. 
The proposed metric is reference-free and training-free, outperforming state-of-the-art methods with a comparable number of model parameters.
For turn-level dialogue evaluations, our experimental results demonstrate that this metric can serve as a generalized alternative to the Negative Log-Likelihood scorer for multi-dimensional evaluation metrics.
We plan to extend our approach to other dialogue evaluation methods and explore its applicability to general text generation problems. We are also interested to see if our measure can improve the factual consistency evaluation for document-grounded dialogue or conversational question answering.
Additionally, we will investigate incorporating our C-PMI-based metric into the fine-tuning process of LLMs.

\section*{Limitations}

While our proposed method demonstrates promising results and outperforms several state-of-the-art techniques, it is important to acknowledge certain limitations.

\begin{itemize}
    \item \textbf{Dependence on pre-trained LLMs:} Our method relies heavily on the pre-trained LLM's quality and the knowledge it has captured. As a result, any biases, inaccuracies, or limitations present in the LLM may directly impact the performance of our evaluation metric.
    
    \item \textbf{Lack of diversity in the dataset:} The FED dataset, which we use for evaluation, is primarily derived from conversations with the Meena and Mitsuku chatbots. Consequently, it is possible that our evaluation might not have better correlation with human ratings for other dialogue systems or more diverse conversational contexts.
    
    \item \textbf{Adaptability to new evaluation dimensions:} Our method currently focuses on eight turn-level metrics. Extending the method to incorporate additional or novel evaluation dimensions might require further investigation and calibration.
    
    \item \textbf{Computational cost:} The current implementation of our approach is around twice as slow as the baseline NLL-based method due to multiple times of the inferences of the language model. The efficiency of the implementation can be improved in the future by re-using the log-likelihood of the dialogue history.
    
    \item \textbf{Subjectivity in human judgments:} Our evaluation metric's correlation with human judgments serves as a key performance indicator. However, human judgments are inherently subjective, which could lead to inconsistencies or discrepancies in the evaluation results.
    
\end{itemize}

Despite these limitations, our proposed method presents a significant step forward in dialogue evaluation, offering a model-agnostic, unreferenced, and training-free approach that captures the human and the system interaction. Future work could address these limitations and explore additional dimensions of evaluation, further refining the method and its applicability across a broader range of dialogue systems and text evaluation systems.
\section*{Ethics Statement}

In this study, we recognize the importance of ethical considerations in natural language processing and dialogue systems research. Acknowledging the potential biases in pre-trained LLMs and human judgments, we advocate for future research to investigate and mitigate these biases in evaluation metrics. We strive for fairness and inclusivity by designing our method to be generalizable and adaptable to various settings. As researchers, we are committed to responsible AI development and contribute to the ongoing discourse on evaluating dialogue systems, enabling the creation of more effective and ethical AI-powered conversational agents. We encourage the research community to continue discussing ethical considerations and promoting transparency in the field.

\section*{Acknowledgements}
We would like to acknowledge support from the Amazon Alexa Prize as part of SocialBot Grand Challenge 5.
% amazon socialbot challenge grant

% Entries for the entire Anthology, followed by custom entries
\bibliography{custom}
\bibliographystyle{acl_natbib}

% \appendix
% \input{700appendix}

% This is a section in the appendix.

\end{document}